\documentclass[10.5pt]{article}
\usepackage[numbers, square]{natbib}

\usepackage{jmlr2e}
\usepackage[dvipsnames]{xcolor}
\usepackage{amsfonts} 
\usepackage{graphicx}
\usepackage{amsfonts}
\usepackage{algorithm}
\usepackage{algorithmicx}
\usepackage{algpseudocode}
\usepackage{array}
\usepackage{amsmath}
\usepackage{booktabs}
\usepackage{siunitx}
\usepackage{bbm}
\usepackage{amssymb}
\usepackage{caption}
\usepackage{subcaption}

\newcommand{\bbeta}{\boldsymbol\beta}

\newcommand{\btheta}{\boldsymbol\theta}

\newcommand{\eeta}{\boldsymbol\eta}
\newcommand{\bmu}{\boldsymbol\mu}
\newcommand{\bgamma}{\boldsymbol\gamma}

\newcommand{\bX}{\boldsymbol X}

\newcommand{\bx}{\boldsymbol x}

\usepackage{url}

\usepackage[dvipsnames]{xcolor}

\long\def\comment#1{}

\jmlrheading{1}{2025}{1-xx}{5/23}{xx/23}{Hlav\'a\v ckov\'a-Schindler, W\"o\ss, Pecorino and Schindler}

\ShortHeadings{}{}

\firstpageno{1}

\begin{document}

\title{Cause or Trigger?
From Philosophy to Causal Modeling}

\author{\name Kate\v{r}ina Hlav\'a\v{c}kov\'{a}-Schindler \email katerina.schindlerova@univie.ac.at \\
       \addr
       %   Data Mining and Machine Learning Research Group\\
       Research Group Data Mining and Machine Learning, \\ Faculty of Computer Science, University of Vienna,
       Vienna, Austria
       \AND
       \name Rainer W\"o\ss \email rainer.woess@univie.ac.at \\
       \addr  Research Group Data Mining and Machine Learning, \\
       Faculty of Computer Science, University of Vienna,
       Vienna, Austria
       \AND
       \name Vera Pecorino \email pecov1800@gmail.com \\
       \addr Department of Physics and Astronomy, University of Catania, Catania, Italy
       \AND
       \name Philip Schindler \email a12315546@unet.univie.ac.at\\
       \addr Faculty of Philosophy, University of Vienna, Vienna, Austria
       }

\editor{}

\maketitle

\vspace{0.5cm}

\begin{abstract}%
Not much has been written about the role of triggers in the literature on causal reasoning, causal
modeling, or philosophy. In this paper, we focus on describing triggers and causes in the metaphysical sense
and on characterizations that differentiate them from each other. We carry out a philosophical analysis of
these differences. From this, we formulate a definition that clearly differentiates triggers from causes and
can be used for causal reasoning in natural sciences. We propose a mathematical model and the Cause-
Trigger algorithm, which, based on given data to observable processes, is able to determine whether a
process is a cause or a trigger of an effect. The possibility to distinguish triggers from causes directly from
data makes the algorithm a useful tool in natural sciences using observational data, but also for real-world
scenarios. For example, knowing the processes that trigger causes of a tropical storm could give politicians
time to develop actions such as evacuation the population. Similarly, knowing the triggers of processes
that cause global warming could help politicians focus on effective actions. We demonstrate our algorithm
on the climatological data of two recent cyclones, Freddy and Zazu. The Cause-Trigger algorithm detects
processes that trigger high wind speed in both storms during their cyclogenesis. The findings obtained
agree with expert knowledge.

\end{abstract}
\vspace{0.3cm}


\section{Introduction}
Causal reasoning has been an inseparable part of every scientific discipline.  Scientists look for causation among studied entities and draw causal conclusions.  More recently, causal modeling  has been  part of the disciplines that use data or observations, such as  climatology, bioinformatics, or computer science.  However, not much has been written about the role of triggers in the literature on causal reasoning, causal modeling, or  philosophy.
Distinguishing a trigger from a cause is important to prevent dangerous situations in climatology, such as flooding, or to prepare anticipatory humanitarian actions.
In climatology or in real-world physical systems, one cannot usually cancel causes. 
 But, for example,  in climatology, in case of a dangerous or short- or long-term extreme situation,  one can  strive  to prevent or temporarily draw away a trigger, accelerating an effect. Short-term extreme situations include for example cyclones or hurricanes, while   long-term  extreme situations involve the the global warming of the Earth's atmosphere. 
Knowing the triggers of these processes could allow scientists in some situations to more directly influence them or at least prepare for the triggered situations.

In the current  impact-based forecasting context, detecting  triggers from past climatological scenarios helps provide decision-makers with the necessary information to know when and where early action should take place. 

One way to distinguish between a cause and a trigger is that as soon as the trigger arrives, the effect  occurs, provided the cause is already there. 
A trigger does not have to be a process which takes place parallel to the causal process;  
it can suddenly occur at some point in time. 
In the world, there are situations when the trigger is clearly separable from the cause.
For example, a rifle has a trigger to allow a projectile to be fired, and water temperature needs to reach a certain point to become ice.

Similarly, in the universe, there are situations where the trigger is clearly separable from the cause. From an astronomical perspective, the ortho-para conversion of molecular hydrogen on amorphous water ice is particularly significant. In this process, the intrinsic magnetic dipole interaction of hydrogen is the cause, and the catalyst, whether it is the ice surface or a nearby magnetic molecule, plays the role of the trigger by providing the magnetic environment needed to perturb the hydrogen molecule, whose transition between the ortho and para forms constitutes the effect.

\begin{figure}[!tbp]
  \begin{subfigure}[b]{0.5\textwidth}
    \includegraphics[width=0.96\textwidth]{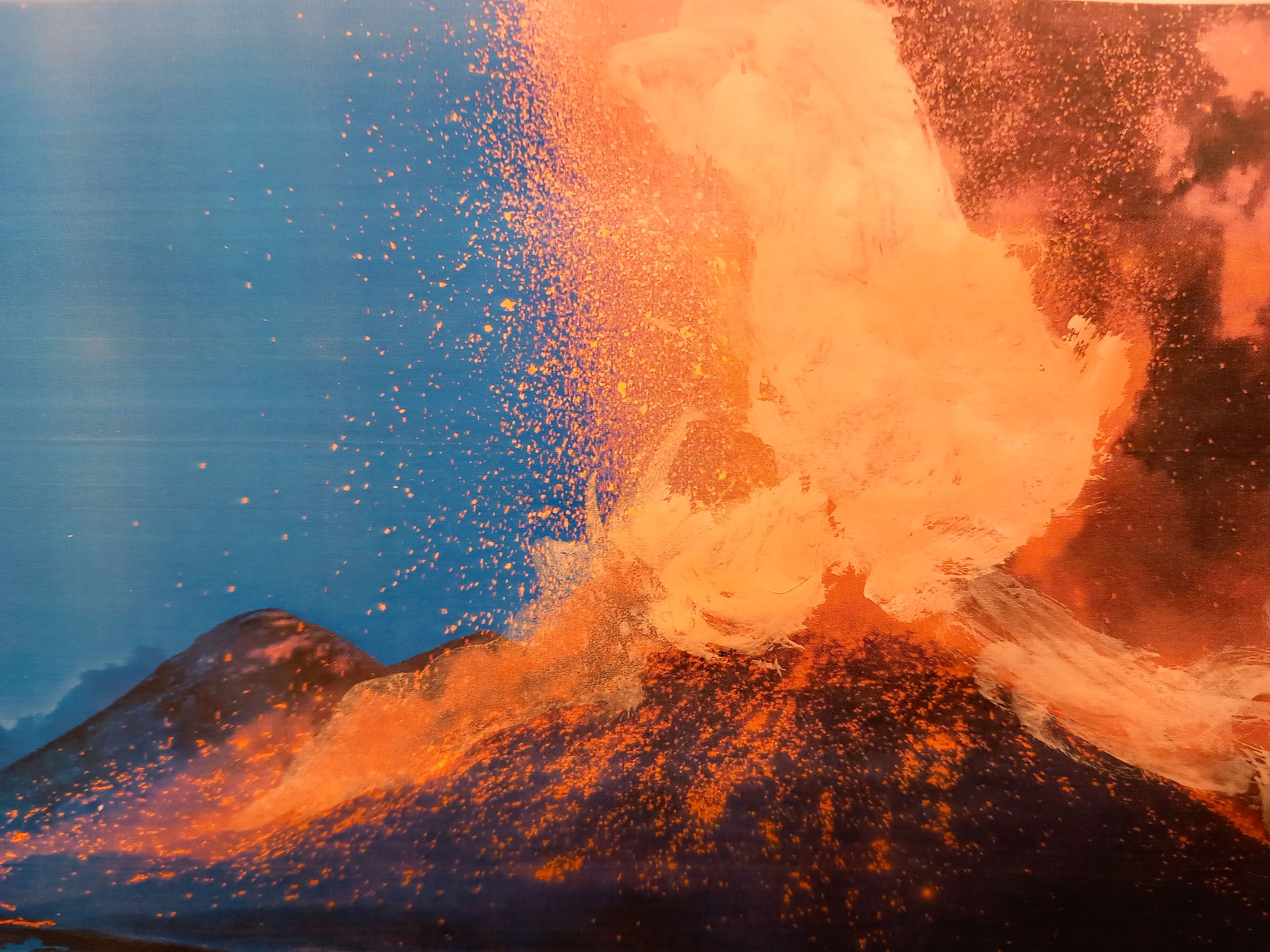}
    \caption{Volcano Etna in February 2021}
    \label{fig:etna}
  \end{subfigure}
  \hfill
  \begin{subfigure}[b]{0.5\textwidth}
    \includegraphics[width=0.96\textwidth]{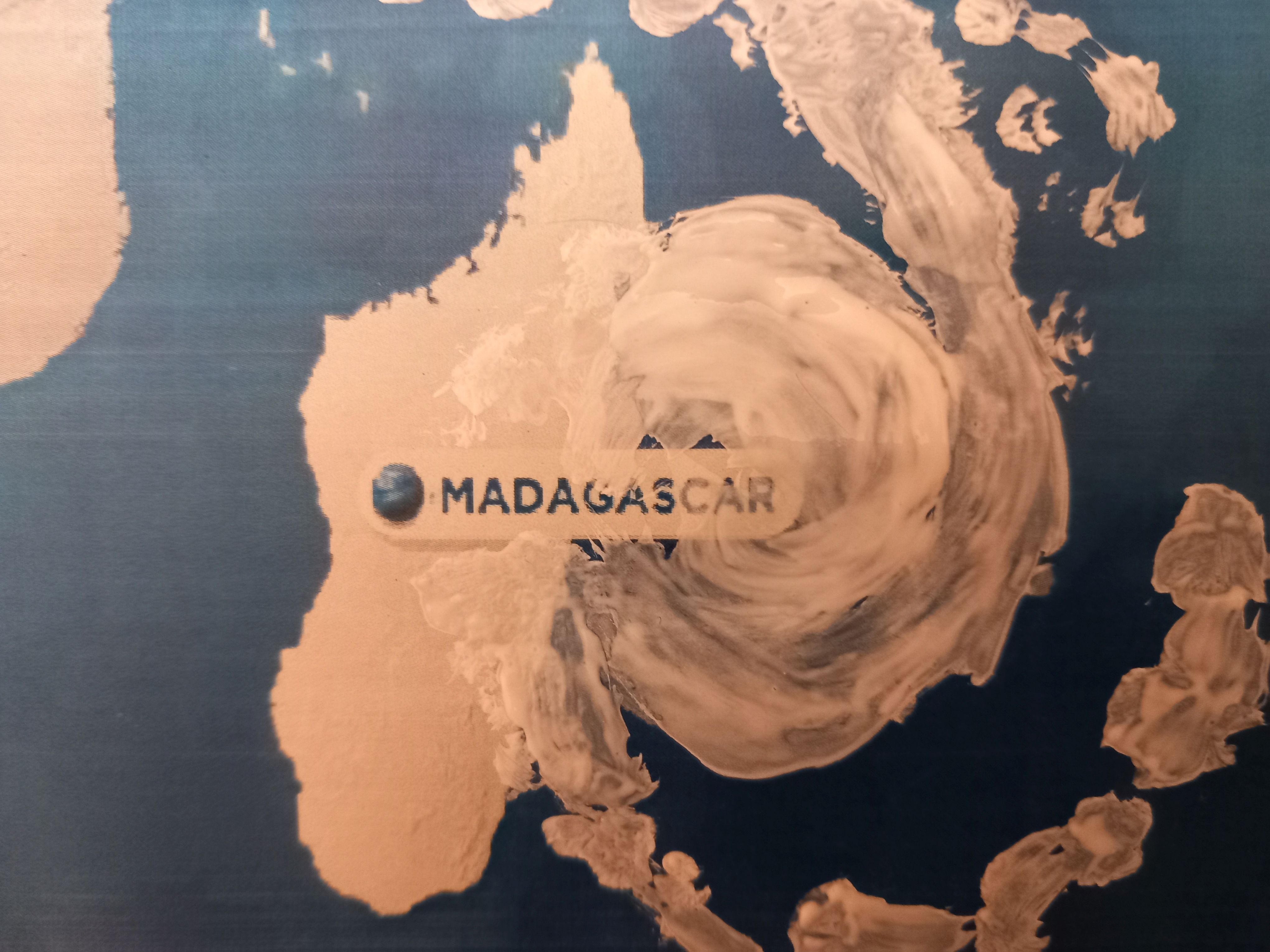}
    \caption{Cyclone Freddy on February 21, 2023}
    \label{fig:freddy}
  \end{subfigure}
  \caption{(a) Volcano Etna in February 2021; (b) Cyclone Freddy approaching Madagascar on February, 21st, 2023.  Drawings made by the first author based on the fotos in public domain.
  }\label{etna}
\end{figure}

During a volcanic eruption, complex interaction  processes are involved, for example magma generation and ascent, gas exsolution  or fracture formation and propagation, \cite{corsaro}, \cite{cas}.  It would be interesting to differentiate between  processes that are causal and those that trigger them. Figure~\ref{fig:etna}  shows  the recent eruption of Mount  Etna  in Sicily from February 2021.

Every realized trigger can trigger (in the sense of its generation) another trigger or  cause. A trigger does not have to be essential for the said effect to occur in general. In many  cases, no trigger is necessary for the relationship between a cause and effect to hold. In  some cases, for a certain phenomenon  to be  a cause of a certain effect,  a trigger may be necessary. At the same time, the same effect might also  be a result of  another cause that does not precede or include a trigger (a chain of causes and triggers).
There can be a plethora of causes and effects happening not just at the same time, but sometimes even causing and affecting each other. In some cases, another variable can help to distinguish: the trigger.
In this paper, we will  use the more general  term  "causal mechanism" which captures both trigger and cause.

Let us look at the following example: Having a weak immune system allows a virus to start a sequence of reactions that almost deterministically lead to an illness. The sequence of reactions in the body depends on the immune system of the host. If the host had a strong immune system, the virus would not trigger the chain of causes that lead to the illness. Is a weak immune system or the virus a cause of the illness? Is the virus a trigger?

In the development of an infectious disease, a trigger seems to be an external variable that influences the causal direction between other variables and comes from its latent form into the visible form. The situation in a human body is very complex, and we are aware that there are other factors involved in the development of a disease. However, this example makes it clear how important it is to know, also in medicine, what is a cause and what is a trigger in the development of a disease.

In this work, we  have the following objectives. Firstly, we  differentiate between trigger and cause in a causal mechanism and  develop definitions which can be used for quantitative causal models in the natural and formal sciences. Secondly, we propose a quantitative causal model and an algorithm that distinguishes causes and their triggers on time series observational data. We demonstrate the plausibility of this algorithm in real-world climatological scenarios; namely, we find triggers in the cyclogenesis of  cyclones Freddy and Zazu. 
Figure~\ref{fig:freddy} illustrates  cyclone Freddy approaching Madagascar on February 21st, 2023.  It would be both interesting and useful to know which climatological variables trigger the high wind speed in the genesis of cyclones or hurricanes.
Let us first focus on the philosophical view that deals with causality and triggers.

\section{Causality in philosophy}\label{Cinphilosophy}

The causality debate in philosophy can be classified into two questions,   metaphysical and epistemic. The metaphysical question concerns the nature of the connection between cause (C) and effect (E): How and by virtue of what does the cause bring about the effect? The epistemic issue concerns the possibility of causal knowledge: How, if at all, can causal knowledge be obtained?

Natural sciences such as physics, mathematics, biology, etc. work more or less with the metaphysical concept, while the social sciences as psychology or sociology utilize the epistemic concept.
Natural sciences describe 
causality (also known as  causation, causal connection or cause-and-effect relationship) as an influence by which one event, process, state, or object (a cause) contributes to the production of another event, process, state, or object (an effect) where the cause is partly responsible for the effect, and the effect is partly   dependent on the cause, as stated in \cite{wisdom1960causation}. 
The effect is always a
change of a body or a system.
The same paper states that 
"... The principle of causation plays an important
role, though not a dominating one. It may be
remarked that this may sound insipid, but if so, it is
not insipid through compromise, but because, if the
author is right, the world is governed by heterogeneous
types of law."
% The following paragraph is from Anupam
Heterogeneity in laws across natural phenomena emphasizes the intricate and multifaceted nature of the universe. Nature exhibits diverse behaviors and patterns governed by various laws and principles. From the microscopic domain of quantum mechanics to the macroscopic scale of cosmology, each aspect of the natural world adheres to its own set of governing principles. This heterogeneity is evident in physical, chemical, biological, and ecological manifestations. While some phenomena  adhere to deterministic laws, others are characterized by stochasticity. Moreover, emergent complex phenomena arising from the nonlinear interactions of system components further contribute to the heterogeneity of natural laws. Therefore, adopting and comprehending this heterogeneity is essential for advancing our understanding of the universe. 
 
There are various conceptions of causal connection (CC) in philosophy.
\cite{krajewski1997energetic}  distinguishes between four philosophical conceptions of a causal connection: 
1. The dynamic
or materialist conception: CC is an action of one body (system) onto
another one, see e.g., \cite{mcginn} and \cite{krajewski1982book}.  2. The voluntarist or spiritualist conception: the genuine
cause is a spiritual being acting consciously with a will, see e.g.,  \cite{oconnor} and \cite{krajewski1982book}.  3. The aprioristic
or logical conception: There is a logical link between cause C and effect E, that is, one can say that  C
is the reason of E, its sufficient or necessary condition, etc., see \cite{krajewski1982book} 4. The
phenomenalistic or positivistic conception: CC is merely a constant succession
of two observed events, see \cite{krajewski1982book}.
The definition of trigger (T) in psychology  resembles  the philosophical conception of cause in terms of 2.
A trigger is described as 
 a stimulus that elicits a reaction, for example,  spider phobia  as described in \cite{peperkorn}. In this sense,  an event (a smell, a figure) could be a trigger for a memory of a past experience and an accompanying state of emotional arousal.

For the topic of our paper,  causal modeling in the broad sense as a metaphysical approach that uses quantitative entities and does not  consider any voluntarist component, we will use the first definition of causal connection,  the dynamic or materialistic conception,  appropriate to processes in natural sciences. 
With this definition, we will reason about the trigger which corresponds to the metaphysical concept, so that it can be used in causal modeling and variables operating in the causal mechanism, i.e. cause, trigger and effect can be determined and quantified.
 
Let us still  summarize the philosophical view on the causal connection from \cite{krajewski1997energetic}. Krajewski argues that there are various kinds of causal connections, namely 
   cause as the supply of energy, cause as a  trigger (releasing factor), and the cause as the supply of structural information.

\subsection{Cause as the supply of energy}
The causal view as the supply of energy follows Robert Mayer, who discovered the law of conservation of energy and explained the CC by means of energy transfer, see e.g., \cite{mittasch1940entwicklung}. Krajewski calls the action that supplies the energy needed for the E energetic cause or the energy cause.

\subsection{Cause as the trigger (releasing factor)}

When an
amount of potential energy  accumulates in a material system, an \emph{impulse}
is usually needed to release this energy. This is another kind of C
which may be called  a triggering cause or trigger cause.
The potential energy may be gravitational (a stone fall  triggers
an avalanche), chemical (a spark triggers a  fire), or nuclear (completion of
the critical mass triggers the explosion of the bomb).
 E  does not have to happen immediately after the trigger. For example, 
 triggering a remote explosion  by  pressing a button initiates
a long sequence of events leading to that explosion. The trigger starts  a chain of energetic causes.
However, the energy in all of them is much lower than the potential energy
which produces the explosion.
 \cite{ostwald1902vorlesungen} distinguished even two kinds of
releasing cause: (a) the total one,  where all the  accumulated energy is released
at once (explosions, avalanches), and (b) where the gradual, regulated one: the energy is gradually released (contraction of a muscle, turning off a tap). In chemistry, one can imagine that  a catalyst is a trigger for a chemical reaction.

Let us illustrate the difference between cause and trigger with the example of tipping water out of  glass.  The cause (increasing level of water) creates the potential for the tipping point to occur; the trigger makes it happen. One way to distinguish between  cause and  trigger is that as soon as the trigger arrives, the tipping point occurs, provided the cause is already there. It is as if the cause was waiting for the trigger to come. 
A trigger without the presence of the cause cannot be effective.  Causes are in most situations  internal, while triggers are external. 
By knowing and identifying the internal causes, we can change the threshold at which the triggers start to matter. 

Consequently, we adopt the standing definition of a trigger from \cite{krajewski1997energetic}, which is characterized by 1. being of lower energy relative to the cause it triggers and 2. preceding the cause on a time scale.
It seems that current philosophy does not address  the problem of triggering.  
This may be due to the fact that some philosophers of science claim that there is only one cause, namely the energetic one, and its release is not considered a cause, see \cite{hartmann1948kausalitat}.

\subsection{Cause as the supply of structural information}
This type of cause, as differentiated by Krajewski can be illustrated by the following example:
the cause of an infectious disease can be the penetration of bacteria or viruses
into the organism. It is not an energetic cause  of the disease, but 
 a triggering cause which is  informational,  since the pathological changes in the tissues 
 bring new  information to the organism.

 In any case, whether we call a trigger  a special form of cause or whether a trigger is a different phenomenon than a cause,  triggers play an important role in causal research.

 \section{Quantification of  cause and trigger among physical processes}

To our best knowledge, we are not aware of any systematical research that focuses on distinguishing between a cause and trigger in physical sciences.
Recently,  the most desirable and topical area for investigating the cause and trigger would be climatological research. For example, one could investigate  the  impact  of  various climatological processes  on global warming.
Currently, international climatological research and related organizations provide a database 
of so called triggers, e.g., in \cite{AnticipationHub}. A trigger is  defined here as  a key component to developing anticipatory action, e.g., evacuation of the population before a predicted cyclone.
%\url{https://www.preventionweb.net/news/trigger-database-live}.
However, triggers  are mainly causes in the sense of the above definition, since no distinction between  trigger and cause is made.

\subsection{Required properties of the cause-trigger model}

From the observations made in Section~\ref{Cinphilosophy} we will consider the notion of trigger, which corresponds to the metaphysical concept. This notion allows that  all three variables involved in a causal mechanism - cause, trigger, and effect -
can be quantified, and the  notion of time can  be used.
 A good  model distinguishing cause and trigger should take into account the following facts:

1. The cause precedes the trigger in time.
 2. Both the cause and the trigger precede the effect.
 3. A trigger without a cause does not increase the change in information (energy) in effect.
 4. In comparison to the increase of information (energy) between cause and effect, the
 considerably lower change of information (energy) by the information (energy) transfer
 is characteristic for a trigger.

If cause and trigger can be distinguished in terms of energy, one must first know what  proportion of energy is the maximum for a trigger to be a trigger and not itself turn into a cause. The found proportion will probably vary for various physical processes. And one must ask whether the values of the trigger variable influence the values of the causal variable.  
We note that this attempt to distinguish  cause from trigger assumes cause and trigger to be of the same quality but varying in quantity of energy (determined relatively not absolutely) causing the effect.

\section{Cause-Trigger algorithm}

In the following, we focus on existing quantitative models that use at least three variables and thus can be used for modeling a cause, trigger, and effect on observational data.
A candidate among existing models for 
  distinguishing between trigger and cause could be the regression moderation model, see e.g., \cite{cohen}.
 Before we test whether a candidate for trigger variable acts in the moderation model as a moderator, we must ensure that it is not a confounding variable 
\footnote{A confounder is a variable that causes both the predictor of interest and the outcome variable.}.
%(A correlation with the predictor and outcome is not sufficient for a variable to be a confounder). 
A moderator  is a variable that influences the strength of the relationship between two
variables.  It varies  the value of the  effect.
A trigger does not affect the causal variable, but affects the strength of the causal variable on the effect. Therefore, a trigger is not a confounder.

Based on the reasoning carried out above in this paper, a good model  distinguishing  the cause and trigger, compared to the moderator model, should  have these properties:

\begin{itemize}
\item[-] Both  cause and trigger precede the effect.
\item[-] The cause precedes the trigger in time.
\item[-] Trigger (similarly to  a moderator) is not necessary for a causal relationship between two variables. It interacts (possibly non-linearly) with the cause.

\item[-] The trigger (similarly to a moderator) without the cause does not increase the change in information (energy) in effect.
\item[-] Cause without a trigger  increases the change of information (energy) in the effect (usually) less than with the trigger.
\item[-] Similarly to a moderator, a trigger affects the strength of a causal relationship. However, in the presence of a trigger,  the strength of a causal relationship is suddenly increased, while for a moderator it  changes gradually.
\item[-] In comparison to the increase in information (energy) between cause and effect, the considerably lower change in information (energy) by the  information (energy) transfer is characteristic of a   trigger (and in fact also of a moderator).
\end{itemize}

\noindent Assume  now that  we have obtained a set of causal variables for a target variable by some causal inference method. We limit ourselves to causal methods working with time series, since we investigate physical processes. The method  used to find them is not essential. 
We only assume that the obtained set of causes and triggers does not contain confounders. 
In the algorithm to distinguish between causes and triggers,  which we propose in the next section, we use  the HMML  method from \cite{hlavackova2020}  to determine the  set of all  potential causes and triggers for a given target, but another causal inference algorithm for time series 
can be used. 
To maintain the fluency of the reading, the HMML method is described in the Appendix.

\begin{algorithm}[H]
\caption{\textbf{Cause-Trigger:} Distinguishing causes and their triggers of a target time series}
 \label{alg6}
\begin{algorithmic}[1]
 		\\  \textbf{Input:}  Time series $(\bx_i^t)  \in  \mathbb{R}, i=1, \dots, p$  in time interval $I= (0,\tau)$  as candidates \\of causes or triggers to  target  $y^t$ in  $I$. ($E(.)$ denotes a mean value).  
   \State \textbf{Output:} The set $C$ of causes of $y^t$ and the set $T$ of triggers  of $y^t$.
   \State $T:= \emptyset$;
\State  Find time subintervals $I_1 =(0,t_1), I_2 =[t_1, t_2)$ of $I$ such that $|E(y^t)|_{I_2} >
|E(y^t)|_{I_1}$. 
\State %\Mag{New:} 
\textbf{If} $I_1$, $I_2$ do not exist, then $C :=$ the set of causal variables on $I$  and \textbf{stop},\\
otherwise go to  step 8.
%\ag{AG: I think $I_1 =(0,t_1)$. $>$ is a typo.}
\State   Find the sets of causal variables $B_1, B_2$  for $y^t$ on $I_1$ and $I_2$, respectively by a causal method.
\State \textbf{If} $|B_2| < 2$,  
 $T := \emptyset$ and \textbf{stop}.
\State Otherwise find all  $x_s \in B_2$ ($\neq y^t$) s.t. $|E(x_s^t)|_{I_2} >
|E(x_s^t)|_{I_1}$.
%where
%$I_s \subset I_2$ is a compact and  of a maximum size in the sense: If a new time value $t+1$ is added to $I_2$, the  mean  of $x_s$ on $I_2$ will be non-increasing in it.\\ 
%where
%$I_s \subset I_2$ is a compact and  of a maximum size in the sense: If a new time value $t+1$ is added to $I_2$, the  mean  of $x_s$ on $I_2$ will be non-increasing in it.
$T := T \cup \{x_s\}$.
%and store interval $I_s$. 
%\State If  $T:= B_2 \setminus B_1 = \emptyset$ then \textbf{stop} and $C:=B_2$:
%\ag{AG: Please elaborate this point.} \Mag{Done.}
%\State \textbf{Otherwise} $\hat{B}$ is a set of potential triggers. Denote $m: = |\hat{B}|$ and find $x_s \in \hat{B}$ s.t.  $|E(x_s)| \ll |E(x_k)|$ on the whole $I$ and add it into set $T$// {\tt potential  triggers have significantly much   smaller energy than causes},  i.e. the energy of a trigger is $\approx \rho \times $ energy of a cause.
\State \textbf{For each} $x_s \in T$ \textbf{do}

\State  Compute $V:= \bX^{ without-trigger} \mathbbm{1}$ for  $t \in I_2$  where  $\mathbbm{1}$ is a  $((m-1) \times d)  \times 1$ matrix of ones.
%all causal strength values obtained by HMML  on  $I_2$
%excluding the $d$ beta-values corresponding to $x_s$. 
%($\top$ is a sign for a matrix transpose.)
\State By the F-test  on Eq. (\ref{granger-moderator2}) and (\ref{granger-moderator})  decide whether $x_s$ is a moderator  of  $V$ 
%(i.e. corresponding to the whole set $\hat{B}$) 
on $\{d+1, \dots, n\} \subset I_2$.
\State If $x_s$ is a moderator,  $x_u := arg_{x_k, k= 1,\dots, m} \max|E(x_k)_{I_1}- E(x_k)_{I_2}|$, is a cause % from $B_2$     
triggered by it. % $x_s$.
\State \textbf{End} // {\tt to For}
\State Return  the sets of pairs $(x_u,x_s)$ with  the causes   from   $C$ and  their triggers from $T$.
\end{algorithmic}\label{algorithm}
\end{algorithm}

\subsection{Details to Cause-Trigger Algorithm}
\noindent  All time series $(\bx_i^t)  \in  \mathbb{R}, i=1, \dots, p$ are standardized.
%Details to Step 9 can be found in Appendix. 
Steps 9 - 13  will be explained in the following.
First, for  a given $d$  %define
%the $(n-d) \times (md)$ dimensional matrix
%\begin{equation}\label{designmatrix}
%\bX^{Lag}_{t,d} = 
%(x_1^{t-d}, \dots, x_1^{t-1}, \dots, x_{m}^{t-%d}, \dots, x_{m}^{t-1})
%\end{equation}
denote ${\bX}$ the $(n - d)\times (m d)$ matrix
constructed from  all causal and trigger variables from $B_2$ together on $I_2$ as in Step 8, where $n$ is the length of interval $I_2$:
  \begin{equation}\label{matrixX}
{\bX} =
\begin{pmatrix}
x^{d}_1   &  \dots  & x_1^1       & \dots  & x^{d}_m    & \dots & x_m^1 \\
x^{d+1}_1 &  \dots  & x_1^2       & \dots  & x^{d+1}_m  &\dots  & x_m^2 \\
%x^{d+2}_1 & x^{d+1}_1 & \mydots  & x_1^3       & \mydots  & x^{d+2}_p  & \mydots &x_p^3 \\
 \vdots   &   \vdots &   \vdots    & \vdots & \vdots     & \vdots & \vdots \\
x^{n-1}_1 &  \dots  & x_1^{n-d-1} &  \dots & x^{n-1}_m  & \dots & x_m^{n-d-1} \\
\end{pmatrix}.
\end{equation}

\noindent Step 12: Assume $x_s$ is a candidate to be tested for a trigger. 
The matrix ${\bX}^{without-trigger}$ is a $(n-d)\times ((m-1) \times d)$ dimensional matrix.
%(i.e. having  $(n-d)$ rows and $((m-1) \times d)$ columns).
The matrix ${\bX}^{without-trigger}$ is a matrix created from $\bX$ so that the submatrix corresponding to 

 \begin{equation}\label{matrix}
\begin{pmatrix}
x^{d}_s   &  \dots  & x_s^1     \\
x^{d+1}_s &  \dots  & x_s^2     \\
%x^{d+2}_1 & x^{d+1}_1 & \mydots  & x_1^3       & \mydots  & x^{d+2}_p  & \mydots &x_p^3 \\
 \vdots   &  \vdots &  \vdots  \\
x^{n-1}_s &  \dots  & x_s^{n-d-1} \\
\end{pmatrix}
\end{equation}

 i.e., the variable $x_s$ which is a candidate to be a trigger and its lagged values, are omitted from ${\bX}$.
 Thus,  $V = \bX^{without-trigger}\hat{{\bbeta}^*}^{\top}$ is a vector of dimension $(n-d) \times 1$, defined in $I_2$, where $|I_2| = n$. 
 Each i-th column of $V$, namely $V^i$  has dimension $(n-d).$
 We define the Granger-moderator equations for variable $x_l$  %on subset of interval $I_2$, concretely 
 for $t=d+1, \dots, n$ as

     \begin{equation}\label{granger-moderator2}
      y^{t} = \gamma_0 + \gamma_1 V^t  + \varepsilon_y^t
\end{equation}
\begin{equation}\label{granger-moderator}
     y^{t} =  \gamma_0  + \gamma_1 V^t  +  \gamma_2 V^t x_s^t + \varepsilon_y^t 
     \end{equation}
and $V^t$ is defined above for $t=d+1, \dots, t$.
%Denote $r = n-d$,  the size of vector $V$.
The variable $V^t$ for a fixed $t$ is a scalar, i. e. it has only one value. Denote

\begin{equation}  
RSS_1= \sum_{t = d+1}^n |y_t  - \hat{\gamma_0} - \hat{\gamma_1} V^t|^2
\mbox{ and } RSS_2= \sum_{t = d+1}^n |y_t  - \hat{\gamma_0} - \hat{\gamma_1} V^t
- \hat{\gamma_2} V^t x_s^t|^2.
\end{equation}

In the Cause-Trigger Algorithm, line 13, the hypothesis $H_0$ to be tested is : $\hat{\gamma_2}$  has a non-significant value in the second regression equation, i. e. Eq.~(\ref{granger-moderator}). This will be tested by the corresponding F-statistic, more details see in Appendix.

\begin{remark}
 Of course, there can be more causal variables triggered by the same trigger variable. 
  We are interested in finding at least one,  and we search for the one which increases its mean value by the trigger at highest, see 
 Step 14 of the algorithm.
 \end{remark}
 
\begin{remark}
The Cause-Trigger Algorithm  has two hyperparameters regarding   the magnitudes  of the differences in  mean values  (in line 5 for mean values  of $y^t$ and in line 10 for mean values of $x_s^t$).
\end{remark}

\section{Detection of triggers in cyclones  by the Cause-Trigger algorithm}

Our objective in this part of the paper is to identify the triggering variables related to the cyclogenesis process.    We stress that we are aware of the complexity of the dynamics of cyclones and that both the selection of causal variables and the selection of pressure levels are not exhaustive.  
Our objective is to demonstrate the ability of the Cause-Trigger Algorithm  to distinguish between  triggering  and causal processes, and we will illustrate it on two cyclones, Freddy and Zazu.

Typically, a combination of warm water, moist air, and converging winds initiates the development of a cyclone. Once these conditions are met, other factors, such as the Coriolis effect and light upper-level winds, help organize and intensify the storm into a full cyclone. Cyclones are low-pressure systems  rotating  clockwise in the Southern Hemisphere, while  systems with counterclockwise rotating winds in the Northern Hemisphere are called hurricanes.

\vspace{0.2cm}

\noindent \textbf{Cyclogenesis}
 Cyclogenesis begins when warm ocean waters destabilize the air, inducing convection and latent heat release. A weak atmospheric disturbance initiates the process, while low wind shear maintains structural coherence. Moderate shear can aid development by aligning convection with rotation, whereas excessive shear disrupts it, inhibiting intensification. The balance between shear and convection governs the evolution of the cyclone, see  \cite{giuliacci2019}.
Cyclones typically form in regions of low atmospheric pressure. The pressure level at which a cyclone starts, can vary but it generally begins when the central pressure drops below 1000 hPa. As the pressure continues to decrease, the cyclone intensifies, leading to stronger winds and more severe weather conditions.
Other factors, such as sea surface temperature, humidity, and others can play crucial roles in the formation of cyclones.
In our experiments, we selected two cyclones with a similar size of the geospatial grid during their cyclogenesis. 

\vspace{0.2cm}

\noindent
\textbf{Cyclone Freddy:}
With a lifetime of over 35 days, Cyclone Freddy emerged as the longest-lived cyclone on record by
\cite{Liu2023}. 
Freddy originated from a tropical low south of the Indonesian archipelago on February 4, 2023, and quickly intensified as it traveled westward across the Indian Ocean.
At its peak intensity, Cyclone Freddy had 10-minute sustained winds of 230 km/h and a central atmospheric pressure of 927 hPa, making it a very intense tropical cyclone, as reported by the \cite{NHC}.

\vspace{0.2cm}

\noindent
\textbf{Cyclone Zazu:}
Cyclone Zazu was a tropical cyclone that occurred in December 2020. It formed over the South Pacific Ocean and affected regions such as Tonga and Niue (Polynesia).  At its peak, Zazu reached maximum sustained winds of 100 km/h (62 mph) and a central pressure of 985 hPa2.
 The cyclone brought strong winds, heavy rainfall, and rough seas to  affected areas, as reported by  \cite{GDACS2025}.

  For both cyclones, one can expect that wind speed is a strong predicting variable of the strength of the cyclone, but our aim is to detect other factors that play a role in its creation. For example, based on expert knowledge in climatology, we expect to find that the direction of wind (especially upward motion perpendicular to the sea level) has a triggering role in cyclogenesis, while other conditions such as a temperature of 26 degrees Celsius, could be assumed but do not further enhance the process in terms of moderation.

%In our experiments with Algorithm~\ref{algorithm},  we aim to distinguish trigger from causal variables of a wind speed during two cyclones which are summarized Table~\ref{tab:example}.

%\begin{figure}
%    \centering
%    \includegraphics[width=0.7\linewidth]{freddy23track.png}
%    \caption{Track of severe tropical cyclone Freddy whilst within the Australian region (times in AWST, UTC +8). Source: Bureau of Meteorology, licensed under Creative Commons Attribution Australia Licence (CC BY). Retrieved from \url{http://www.bom.gov.au} \Mag{Vera, can you find a similar copyright free figure? Otherwise we must omit it. We anyway do not mention this figure in the text.}}
%    \label{fig:enter-label}
%\end{figure}

%\begin{figure}
%    \centering
%    \includegraphics[width=0.6\linewidth]{Screenshot 2024-10-16 093647.png}
%    \includegraphics[width=0.6\linewidth]{Screenshot 2024-10-16 093745.png}
 %   \caption{06/02/2024 \Mag{Vera, where is the figure from? We must write here its copyright.}}
%    \label{fig:enter-label2}
%\end{figure}

%Satellite Cyclone's track at \url{https://zoom.earth/storms/freddy-2023/#map=satellite-hd }
%{\uline{Freddy satellite tracking}}

\section{Experiments}

\subsection{Data set and variables}\label{dataset}

We used the ERA5 dataset from \cite{hersbach2020} to investigate the potential cause and trigger variables in the context of the cyclone genesis of  Freddy (2023) and Zazu (2020).  The dataset is a state-of-the-art reanalysis from the European Center for Medium-Range Weather Forecasts (ECWMF), which combines model data with observations to provide hourly estimates of atmospheric, ocean-wave and land-surface quantities. The gridded data are available on a spatial resolution between 25-30 km. For our experiments with both cyclones, we used single-level and pressure-based measurements located  within an  approximate 100 km radius around the location of a cyclone's eye at the time of classification or "genesis". For Cyclone Freddy, this results in a $8 \times 8$ grid containing 64 locations within a circumscribed circle within the radius described above.
For Cyclone Zazu, a 9x9 grid of 81 locations 
within a radius of approximately 125 km around the eye was used.

%For cyclone Catarina.... 

 \begin{table}[h!]
\centering
\begin{tabular}{c|c|c|c}
\hline
Name & Area & Cyclone lifetime &  Interval $I$ in Algorithm \\  \hline
Freddy   & South Indian Ocean   & 04.02.23-14.03.23   &  04.02.23 00:00:00 - 07.02.23 12:00:00\\   \hline
%Herold   & South-West Indian Ocean   & 13.03.20-20.03.20   & Data 10  & Data 11  & Data 12  \\ 
Zazu  & South Pacific Ocean  & 13.12.20-16.12.20  & 11.12.20 12:00:00 - 15.12.20 12:00:00  %\\
%\hline
%Caterina  & South Atlantic Ocean  & 12.03.04-28.03.04  & Data 22  & Data 23  & Data 24  
\\ 
%\hline
%Data 25  & Data 26  & Data 27  & Data 28  & Data 29  & Data 30  \\ 
\hline
\end{tabular}
\vspace{0.2cm}
\caption{Characteristics of Freddy and Zazu: areas of occurrence, durations, and key meteorological parameters such \\as storm-to-cyclone transition timestamp along with the corresponding wind speeds and central pressure values. }
\label{tab:example}
\end{table}

The time intervals in which the cyclogenesis developed for each cyclone can be found in Table~\ref{tab:example} and are obtained from 
\cite{ZoomEarth2023}.
 %  \Mag{we could provide here the link to the web cyclone tracker we used to select the time intervals}.

We also stress  here  that our goal was not to determine the  set of all causal variables influencing  cyclone formation. Our 
 goal is to demonstrate the ability of Cause-Trigger Algorithm to distinguish trigger and cause variables under a given set of causal variables.
 
 Keeping this limitation in mind, we select variables, that based on  expert knowledge,  are  relevant to  cyclone dynamics. 
 Instead of the likely influential variable vertical velocity (w),
we construct a special variable $sin(wd)$ where $wd$ is the wind direction. The function sin is a trigonometric function that has its maximum value at 90 degrees or $\pi/2$, i.e. 
$\sin(90 \deg) = \sin(\pi/2) = 1$. This variable models the occurrence of wind speed in the perpendicular direction. Its highest value, 1, is achieved if  the wind direction is perpendicular to the Earth's surface. It is well known that the increasing occurrence of perpendicular wind direction accelerates the wind speed before and during formation of a cyclone, see e.g., \cite{zheng}. 
We use the following set of variables (with their units in brackets) as input for the Cause-Trigger Algorithm:

\begin{enumerate}
    \item Divergence (d) [$s^{-1}$]: The rate at which air spreads out from a given point, influencing cyclone development.
    \item Geopotential (z) [$m^2s^{-2}$]:  
    The gravitational potential energy per unit mass, related to atmospheric pressure levels.
    \item Ozone mass mixing ratio (o3) [$kg kg^{-1}$]: The concentration of ozone in the atmosphere, which affects radiation and temperature.
    \item Potential vorticity (pv) [$K m^2 kg^{-1}s^{-1}$]: The measure of the rotation and stability of an air parcel, crucial for cyclone dynamics.
    \item Relative humidity (r) [\%]: The 
    ratio of actual to maximum possible water vapor in the air, which influences  cloud formation.
    \item Vertical velocity (w) [$Pa s^{-1}$]:
    The speed of air movement in the vertical direction, critical for convection and cyclogenesis.
    \item Temperature (t) [$K$]:
    The atmospheric temperature, fundamental for thermal gradients and cyclone intensity.  
    \item Wind speed (ws) [$ms^{-1}$]:
    The magnitude of the wind velocity, affecting energy transfer and storm development.  
    \item Wind direction (wd) [$^\circ$]:
    The orientation of the wind flow, essential for tracking storm movement and structure.
    \item Sine of the wind direction (sin(wd)) [a value from [-1,1]]: Mathematical transformation of the wind direction.
    %useful for vector decomposition.    
\end{enumerate}

\vspace{0.2cm}

The ozone mass mixing ratio (o3) is among the possible causal variables, 
as it influences air mass movement during cyclogenesis. 
In theory, the ozone  mass mixing ratio could locally impact the movement of air masses during cyclogenesis, but in an indirect and secondary way compared to the main dynamical and thermodynamical drivers of cyclone formation.
More details on possible 
mechanisms of influence of o3 and  other selected variables can be found in  Appendix~\ref{influence_mechanisms}.
We consider the above variables  under a given atmospheric pressure value. 
%Maps of pressure levels are used to identify the locations of low and high pressure weather systems, including  cyclones.
We do not take pressure level as a separate variable  but consider  the above variables under three different air pressure values (pressure levels), namely 
500hPa, 700hPa and 975hPa. 
 Each of these pressure levels corresponds to an approximate height above  sea level with a specific dynamics of a cyclone: 

\noindent     500 hPa ($\approx$ 5,500 m)  \textit{upper-level troposphere}: 
    This level governs large-scale steering flows and mid-tropospheric vorticity maxima. It influences vertical motion through divergence and convergence patterns. \\
    %This level is also useful for diagnosing Rossby wave interactions and upper-level forcing mechanisms. \\
     700 hPa ($\approx$ 3,000 m)  \textit{mid-level troposphere}:  
    At this level, the transition between surface dynamics and upper-air influence takes place. It is critical for identifying vertical velocity patterns, moisture transport, and mid-level vorticity changes. Relative humidity and atmospheric stability at this level influence deep convection.\\
     975 hPa ($\approx$ 600 m)  \textit{near-surface level}:  
    This level captures where surface convergence, temperature gradients, and humidity create instability. Its proximity to the surface makes it suitable for detecting early signs of temperature contrasts and moisture build-up that  are typically observed during the onset of cyclogenesis.
%\Mag{{\tt The previous sentence is dangerous. The reviewer will ask us, why we did not take warm-core anomalies baroclinic zones or other variables measuring these into the set of variables so that our algorithm could detect it as a trigger. Write it differently. -KATERINAAAAA is it now good?}}
We selected these  pressure levels, and for each of them, we used  the 2D grid approximation of 64 locations to visualize the cyclone dynamics in 3D space. More details on the visualization will be given in Sections~\ref{2D_grids} and~\ref{3D-grids}.
%\subsection{Spatio-temporal grid for three pressure levels}
%\section{Results}
By analyzing these three pressure levels, we can explore how surface disturbances, mid-level convection, and upper-level dynamics are interconnected during the development of cyclonic systems. 
%\Mag{{\tt Again, written dangerously. Use the word causal only when we talk about our selected (potentially) causal variables, i.e. those 10 above. Rephrase what you want to say here. In fact there relations could be only correlations, or co-occurrences.KATERINAAAAA is it now good?}}
This multilevel approach is essential for understanding the processes that drive cyclone formation and intensification.

%We excluded from the potentially causal variables also geopotential (z), since  it implicitly depends on the height, which is given by a pressure level.

\subsection{Cause-Trigger algorithm in 2D spatio-temporal grids}\label{2D_grids}

The experiments for each cyclone were performed in the following way.
In three time intervals - clearly before, during and after after a cyclone - we selected one location in the grid where we defined potential causes and triggers affecting wind speed. 
A temperature of 26 degrees Celsius is a boundary condition for considering a time interval of a cyclone.
%\Mag{Rainer, describe it, how you did it at Zazu.}
%\Rainer{the same steps are performed for all cyclones}

%\Mag{Katerina on Feb 24: Rainer, comment please the figures below  in details, as well as refer to Figure 3}.

\begin{figure}[ht] 
    \centering 
    \includegraphics[width=0.8\textwidth]{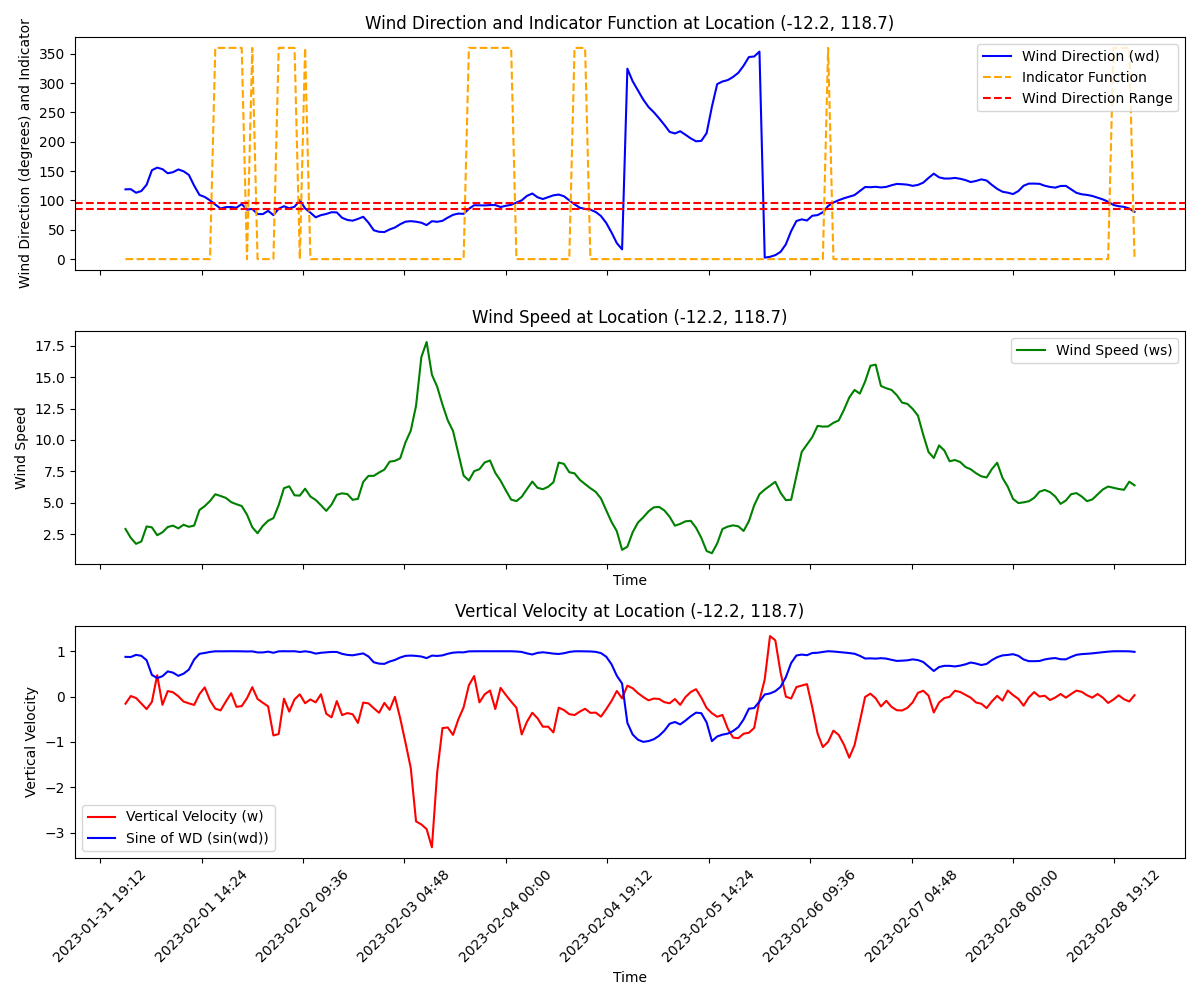} 
    \caption{Example of wind-related time-series at one of the 64 locations of Freddy. 
    The goal of these experiments was to find out in which interval a good separation by detecting the maximal difference of means index can be obtained.}
    \label{fig:timeseries-example-700}
\end{figure}

\begin{figure}[!tbp]
  \vspace{-0.5cm}
\begin{center}
\hspace{1.4cm}
\begin{subfigure}[b]{0.26\textwidth}
\includegraphics[width=\textwidth]{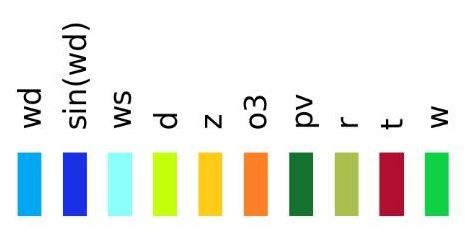}
 \end{subfigure}
 \end{center}
 \hspace{-0.4cm}
  \begin{subfigure}[b]{0.65\textwidth}
\includegraphics[width=\textwidth]%{Freddy_three_levels.png}
    {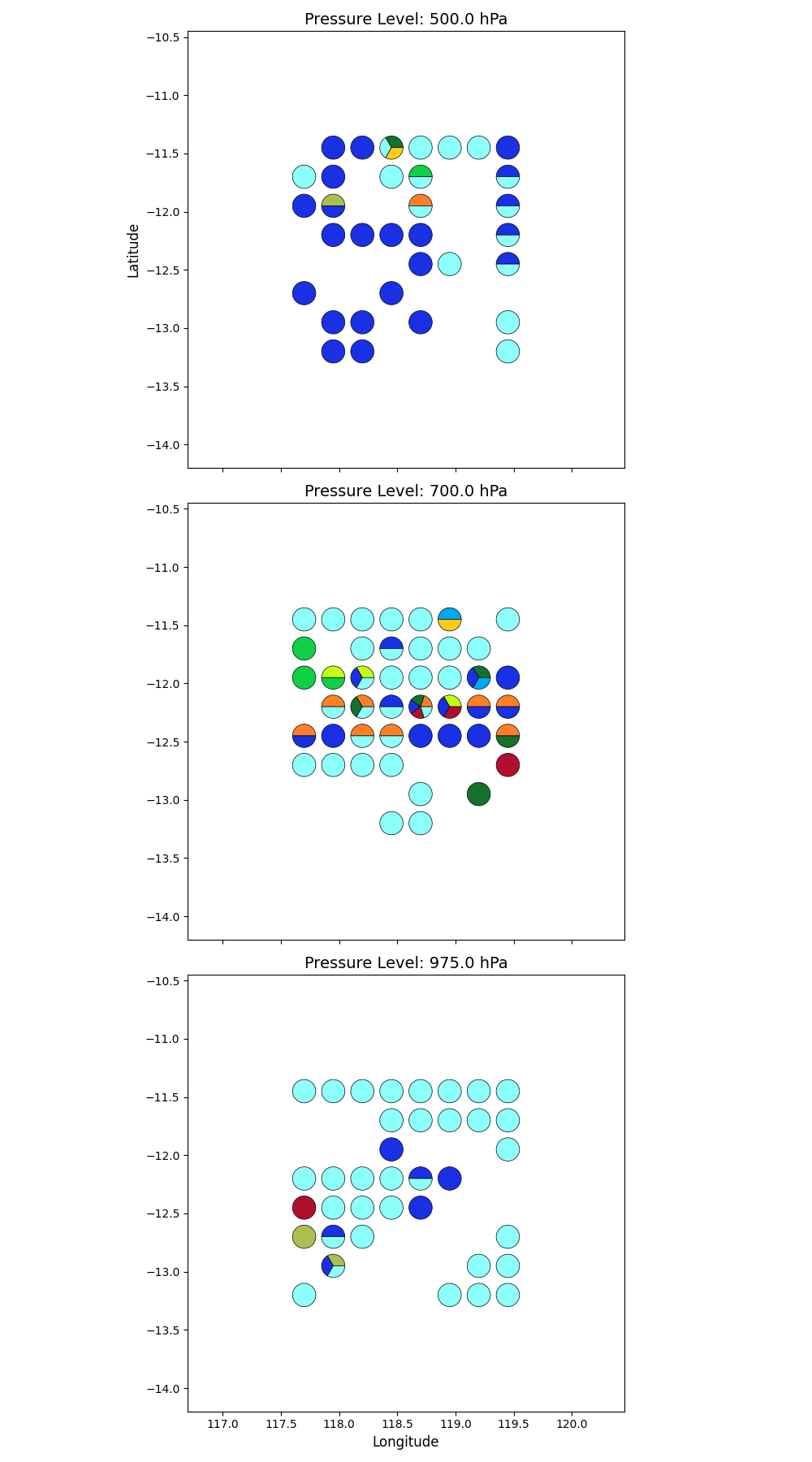}
    \caption{Freddy.}
    \label{fig:f1_2}
  \end{subfigure}
%  \hfill
  \hspace{-2.4cm} 
  \begin{subfigure}[b]{0.65\textwidth}
\includegraphics[width=\textwidth]{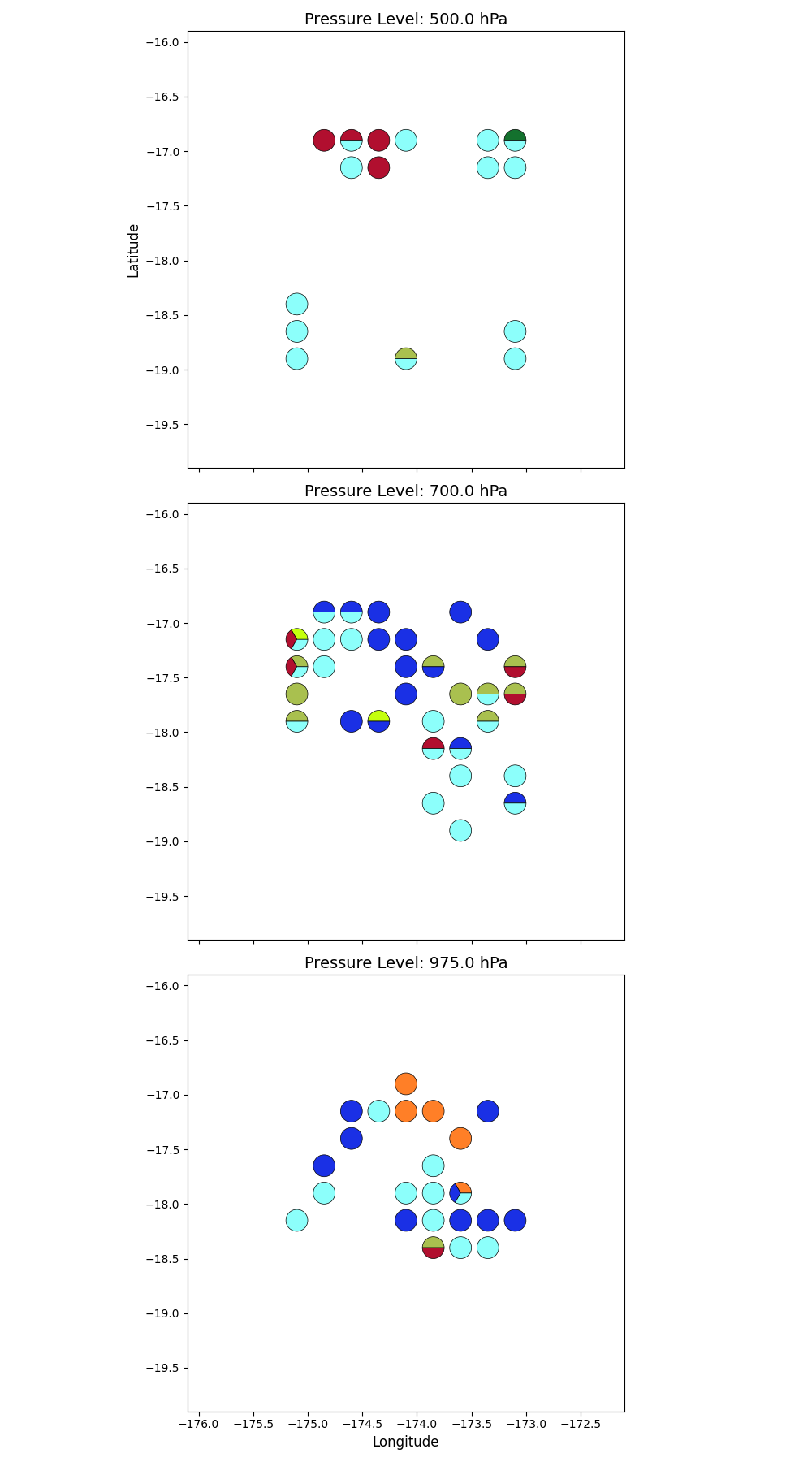}
    \caption{Zazu.}
    \label{fig:f2_2}
  \end{subfigure}
  \caption{Triggering variables detected by Cause-Trigger algorithm per location and pressure-level for both cyclones. Longitude and latitude are shown on the x- and y-axis, respectively. Each color corresponds to a triggering variable.}\label{freddy_zazu}
\end{figure}

%Next steps: 
%3 datasets (single levels, pressure based , combined) -

%\Mag{Rainer: Describe in words how you did the experiments. You can use also your Figure 2 of the 8x8 grid in P1 is an illustrative figure.}
We obtained geographic locations and times of occurrence for both cyclones through Zoom Earth, accessed on
January 25, 2025.
First, we converted the NetCDF4 data, acquired from \cite{hersbach2023}
%\url{https://cds.climate.copernicus.eu/datasets/reanalysis-era5-pressure-levels?tab=download}, 
at the respective times and locations to a .csv file. Then we derived the additional variables from the u- and v- wind components. 
We obtained a dataset containing time series for all variables listed in Section~\ref{dataset} during the time intervals of both cyclones Freddy and Zazu (Table~\ref{tab:example}), at a grid of locations around their respective eyes. This is done for all three pressure levels, namely 500, 700 and 975 hPa.

Each set of variables corresponding to a cyclone, at a geographic location (longitude, latitude) and a pressure level, was individually standardized using the function StandardScaler from the Python library sklearn. %On this data,
The time lag of the time series was selected by evaluating the VAR model using Akaike Information Criterion and the distribution fitting was done  by  the Kolmogorov-Smirnov Test. To construct the subintervals $I_1$ and $I_2$, we iterated over all possible indices to separate the data into two intervals and selected the index that maximizes the difference in mean wind speed for the resulting intervals. Furthermore, we used a parameter to constrain the size of the interval $I_2$ (a minimum of 30 samples). If this step is not taken, there could be very few samples in the interval $I_2$, making it difficult to apply a causal inference algorithm  for time series (in our case HMML) to obtain  potential triggers.

The results of experiments  in a spatial grid for three pressure levels and cyclones Freddy and Zazu are illustrated in Figure~\ref{freddy_zazu}.
Both subfigures (a) and (b) show that, disregarding wind speed itself (light blue), the sine of the wind direction (dark blue) is the most frequently detected 
triggering variable of wind speed in some locations of the cyclones.  Regarding Freddy,  this can be traced back to a relation visible in Figure \ref{fig:timeseries-example-700}, 
where, when the eye of the cyclone, in which it is calm, moves through a point in the grid (at time of February 6) the wind direction slowly begins to turn perpendicular (90 degrees) once more, increasing the value of sin(wd) and the wind speed to their maximal values. 
We can also observe  both for Freddy and Zazu that locations where sin(wd) were found seem to show a counter-clockwise pattern with increasing the pressure levels. This can be explained by physical phenomena typical for  cyclones in the Southern Hemisphere.

We provide an output of causal and triggering variables for both cyclones in the grid location and at each pressure level, as well as a Python code for the Cause-Trigger algorithm under 
%\url{https://zenodo.org/records/15016325}.
\url{https://zenodo.org/records/15109084}.
Concretely, in the csv file (see the supplementary material), the first column lists  the  selected causes and the second column  lists the selected triggers. The i-th values in each list correspond to each other, forming the selected pairs. One 
can observe that the most common pairs for Freddy are [ws,ws] and [ws,sind(wd)]. For Zazu,  the most common output of the Algorithm Cause-Trigger is a combination of ws and sin(wd).

\subsection{Spatio-temporal grids for three pressure levels for Freddy and Zazu and their physical interpretation}\label{3D-grids}

\begin{figure}[!tbp]
%\hspace{-1cm}
\hspace{1.8cm}
  \begin{subfigure}[b]{\textwidth}
\includegraphics[width=0.87\textwidth,height=0.39\textheight]{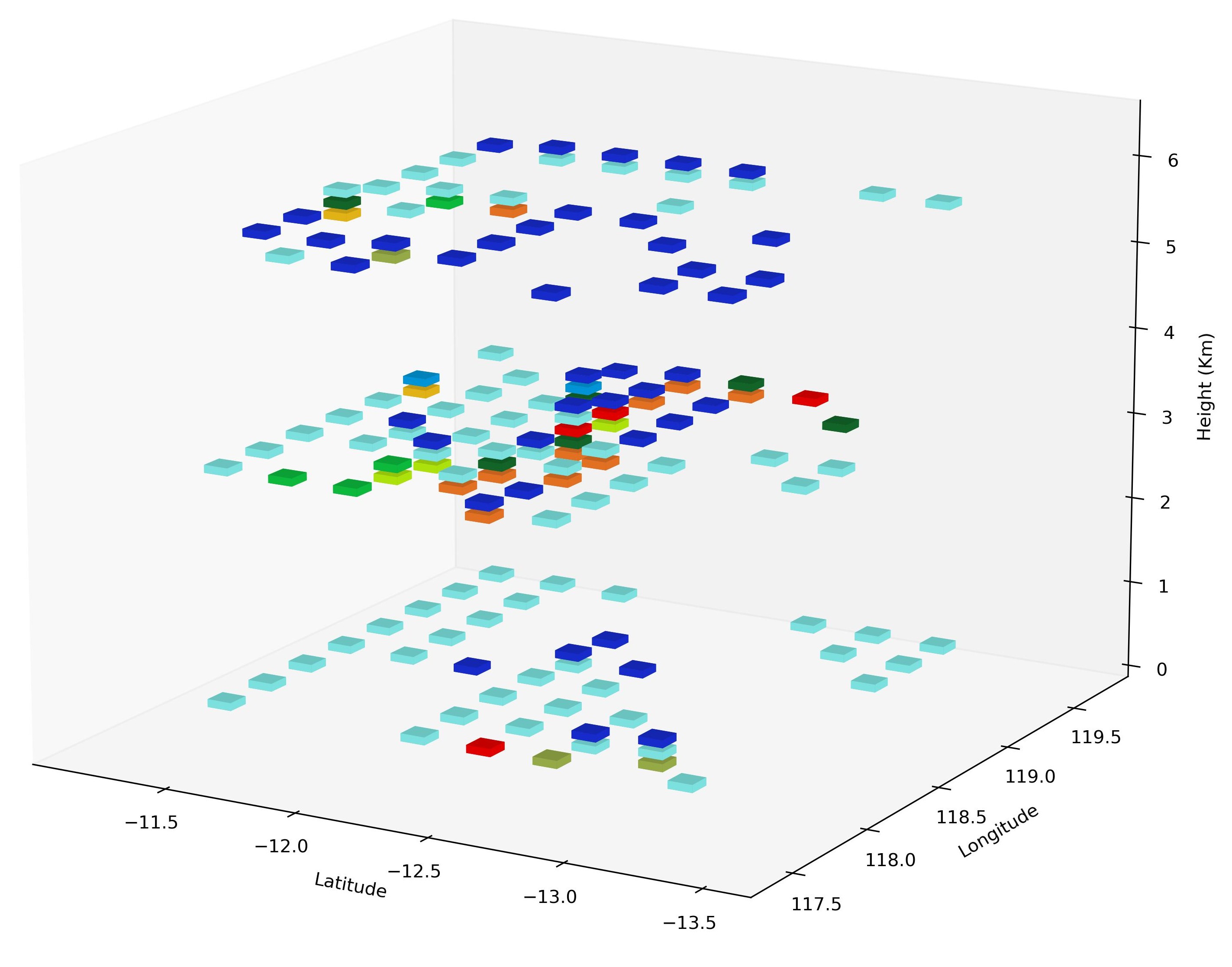}
    \caption{Freddy}
    \label{fig:f1}
  \end{subfigure}
%  \hfill
%  \hspace{-2.3cm} 
  \begin{subfigure}[b]{\textwidth}
  \hspace{1.8cm}
\includegraphics[width=0.87\textwidth,height=0.39\textheight]{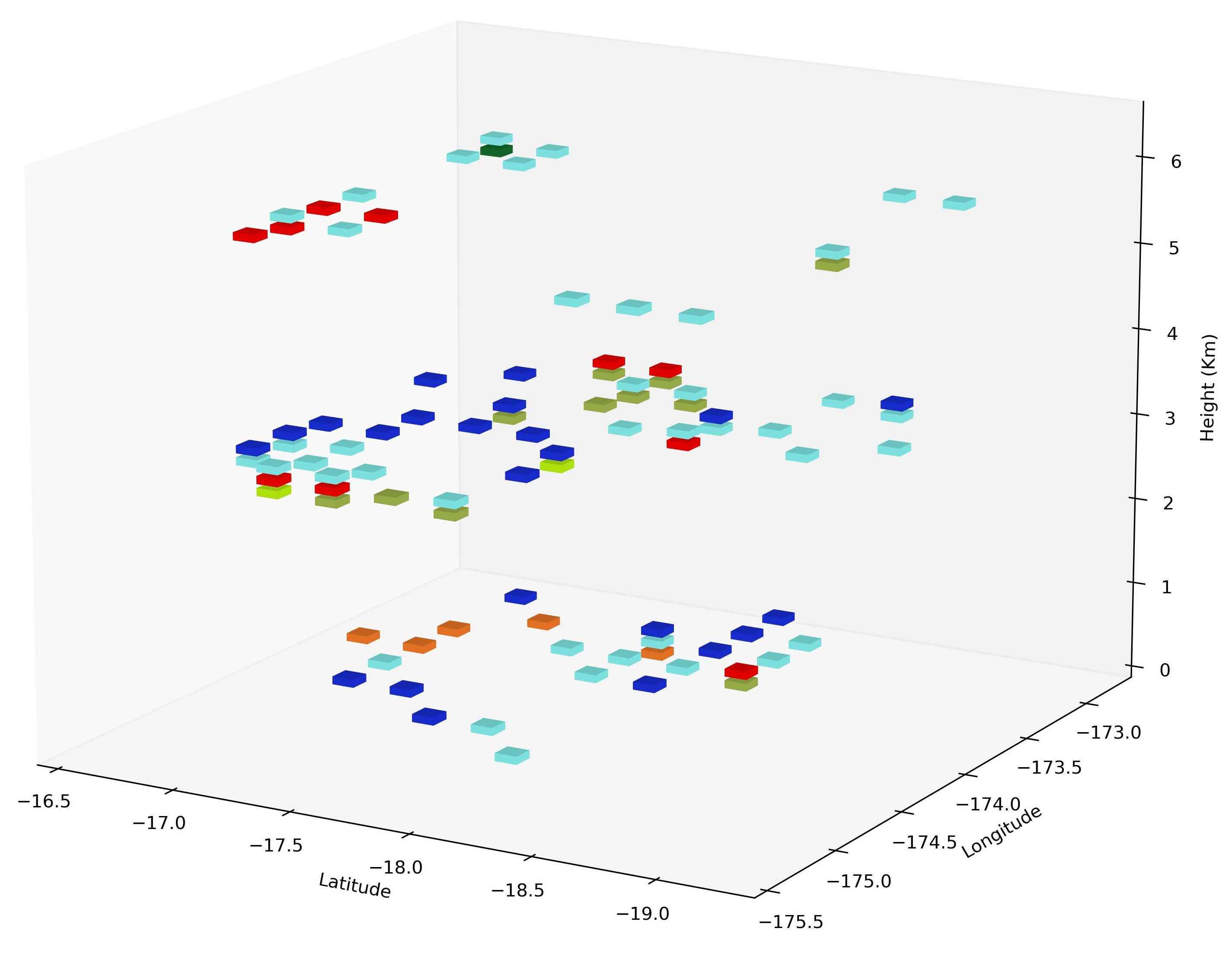}
  \vspace{-0.5cm}
\begin{center}
\hspace{0.9cm}
\begin{subfigure}[b]{0.25\textwidth}
\includegraphics[width=\textwidth]{legend_rotated.jpg}
 \end{subfigure}
 \end{center}
    \caption{Zazu}
    \label{fig:f2}
  \end{subfigure}
  \caption{3D plots of cyclones Freddy and Zazu:  Longitude and latitude are shown on the x- and y-axis, respectively. The z- axis displays the height in km  with respect to the sea surface level. We present only location where triggering variables are active. Each colored cube corresponds to a triggering variable.} \label{fig:3D_both}
\end{figure}

 Cyclone Freddy  is an exceptionally long-lived and dynamically intense system, and Cyclone Zazu is a short-lived system influenced by nearby atmospheric conditions. Using a data-driven causal inference framework, we identified triggering variables on the day each system was first classified as a tropical cyclone. Each spatial site could be simultaneously linked to multiple triggering variables, capturing the multivariate nature of cyclogenesis.
Figure~\ref{fig:3D_both} illustrates the triggering variables detected by the Cause-Trigger algorithm per location and pressure level for both cyclones from Figure~\ref{freddy_zazu} in three-dimensional settings.

Cyclone Freddy exhibited a broad set of triggering variables distributed across all analyzed atmospheric levels (975 hPa, 700 hPa, and 500 hPa), indicating a vertically coherent structure already in place at the time of classification. Wind speed (ws) and  sine of wind direction (sin(wd)) were the most frequent triggers at all levels, particularly at 700 hPa and 500 hPa. These are physically linked to convergence, rotational flow, and upper-level ventilation, see \cite{TangEmanuel2012}, suggesting that Freddy developed strong dynamic organization early on. Potential vorticity (pv) and vertical velocity (w), especially at 700 and 500 hPa, reinforce this picture by indicating enhanced convective activity and barotropic coupling, see \cite{ToryMontgomery2006, MontgomeryEnagonio:1998}. Ozone (o3) also emerged as a significant trigger at 700 hPa and marginally at 500 hPa, a rare finding for tropical systems. Its presence may be indicative of stratosphere–troposphere exchange processes that can influence upper-level thermodynamics, see \cite{HighwoodHoskins:1998, SprengerWernli:2003}.

In contrast to Freddy, Cyclone Zazu showed most triggers at 975 hPa and 700 hPa. Wind speed and sin(wd) were again prevalent, consistent with early-stage cyclonic circulation near the surface. Interestingly, ozone appeared as a significant trigger at 975 hPa, which is a non-typical feature. We hypothesize that this anomaly may be due to the influence of severe tropical Cyclone Yasa, active in the same basin, potentially modifying the near-surface ozone through large-scale subsidence or horizontal transport, see \cite{SchreckMolinari:2011}. At 700 hPa, relative humidity (r), temperature (t), and divergence (d) were also identified as triggers, pointing to a preconditioned environment favoring shallow convection and surface convergence.

The occurrence of certain variables, such as ws and sin(wd), as triggers, despite their causal physical role in isolation, suggests that causality in this context is highly conditional. These variables may act as effective triggers only when some favorable conditions are fulfilled, such as elevated mid-level moisture, existing vorticity, or instability. This reflects the inherently multivariate and nonlinear nature of cyclogenesis, where multiple interacting processes are required to initiate the  development of a cyclone, see \cite{RitchieHolland:1999, marenco2018}. 

While most of the identified variables align with established physical drivers, such as humidity, vertical motion, and vorticity, our results highlight that even weaker or more ambiguous indicators, such as directional components or ozone, can acquire triggering relevance depending on the surrounding atmospheric state. In particular, ozone can serve as a tracer for dynamical interactions between tropospheric and stratospheric layers, or as an indirect proxy  for environmental organizations.

This  spatial analysis also reveals differences in the maturity and depth of the two cyclones at genesis. Freddy’s multi-level triggering structure is consistent with a deeply developed, internally driven system, whereas Zazu appears to be more modulated by its environment and neighboring systems. These findings can be supported by the established literature on mid-tropospheric moisture, see \cite{marenco2018}, low wind shear, see \cite{TangEmanuel2012}, and potential vorticity anomalies, see \cite{ToryMontgomery2006}.

%\textbf{Limitations of the experiments  and possible  future extensions}\\
 Limitations: Our experimental analysis  was limited to two cyclones of similar size and ocean basin, which may limit the generalization of  the results. Our future work on cyclones or hurricanes will explore larger samples across multiple regions.

\section{Discussion and conclusions}

Based on the philosophical analysis in this paper,  we formulated a definition that clearly differentiates trigger from cause and can
be used for causal reasoning in natural sciences. We proposed a mathematical model and the Cause-Trigger
algorithm which, based on given data of observable processes, is able to determine whether a process is a
cause or a trigger of an effect. We provide  Python code for this algorithm. We demonstrated the plausibility and practicality of this algorithm on  two cyclones. The algorithm distinguished between causes and triggers of  high wind speed during cyclogenesis.
There are some limitations of the Cause-Trigger algorithm and directions for future work. 
 The triggering variables, which can be detected by the algorithm, are continuous processes rather than instantaneous events.
Also, our experimental analysis of the cyclones was constrained by the discrete spatial resolution of the selected pressure levels, which are vertically distant and do not provide a continuous depiction of the atmospheric column. These constraints can   distort the spatial continuity and duration of the triggering processes.
Regarding the graphical output of the Cause-Trigger algorithm for cyclones, the 
pies in 2D and columns in 3D  illustrate only the  occurrence of the triggering variable.
Future  work could visualize the relative magnitudes of all triggering variables in a location by   proportions in the pies in 2D or by column height in 3D.
Other future work could  apply the Cause-Trigger algorithm together with an initial  causal method which allows variable lag for each variable.

In conclusion, and to the best of our knowledge,   the Cause-Trigger algorithm is  the first data-driven algorithm  distinguishing between causes and triggers. Its
 applicability  is broad. It can be applied 
in  scientific disciplines using  temporal data measurements of continuous processes, such as in physics or chemistry. 
The Cause-Trigger algorithm offers a useful tool to create new hypotheses about the dynamics of natural processes based on observed data. These hypotheses can not only be of  scientific value but also of societal impact. For example,   knowing the concrete trigger of cyclone development could enable politicians to develop actions such as evacuating populations early enough before a predicted cyclone. 
 Similarly, understanding the triggers of processes causing global warming could help politicians focus on effective actions.

\vspace{0.4cm}

\noindent \textbf{Acknowledgements:}
We  thank  to  Dr. Anupam Ghosh from the Czech Academy of Sciences  for his contribution to the text on the heterogeneity of laws in natural phenomena.  %\Mag{Vera Pecorino acknowledges financial support from ....}

\vspace{0.4cm}

\noindent \textbf{Data availability statement:}
 The dataset, which is a subset of ERA5, and code version as of 30.3.2025 are published on Zenodo: \url{https://zenodo.org/records/15109084}. The full ERA5
dataset is available from Copernicus Climate Data Store: \url{https://cds.climate.copernicus.eu/datasets}.

\vspace{0.4cm}

\noindent \textbf{Author contributions:} All authors have contributed intellectually to the project, and to the drafting of the
manuscript.

\vspace{0.4cm}

\noindent \textbf{Conflict of interest:} The authors state no conflict of interest.

\vskip 0.2in
\bibliography{sample}
%\section*{References}

%\vspace{1cm}

\newpage
\noindent {\Large \textbf{Appendix}}

\vspace{0.3cm}

\noindent  {\large \textbf{Heterogeneous graphical Granger model and its estimation by the HMML method}}
%\Mag{I would say this section is now done.}

\vspace{0.3cm}

Granger causality, introduced by \cite{granger1969investigating} to distinguish between  cause and effect,  can be extended to  the multivariate case, i.e. for $p > 2$  time series and  model order $d \geq 1$, which is   a time lag of   past lagged observations included in the model. The model order can be determined via  information theoretic criteria such as the Bayesian or Akaike information criterion.
For $p$ time-series $\bx_1,..,\bx_p$  the vector auto-regressive (VAR) model is: 

\begin{equation}\label{VAR_for_Granger}
x_i^t =\bX^{Lag}_{t,d}\bbeta_i' + \epsilon_i^t
\end{equation}

where $\bX^{Lag}_{t,d} = (x_1^{t-d},..,x_1^{t-1},..,x_p^{t-d},..,x_p^{t-1})$. $\bbeta_i'$ is the transposition of the matrix $\bbeta_i$ of the regression coefficients and $\epsilon^t$ is the error (see \cite{behzadi2019granger}). 
One can state that \emph{the time-series $\bx_j$ Granger-causes the time-series $\bx_i$ for lag $d$ and denote 
$\bx_j \to \bx_i$, for $i, j = 1, \dots, p$  if and only if at least one of the $d$ coefficients in row $j$ of $\bbeta_i$ is non-zero}. Thus, to detect causal relations, the coefficients of the VAR model are to be determined.

Multivariate Granger causality, among time series from Eq.~(\ref{VAR_for_Granger}),  as a special case of graphical causal models,  assumes that random error time series follow Gaussian distributions with zero mean and constant deviation. 
This assumption might be violated in many applications and a graphical Granger model can infer inaccurate or spurious causal relations. 
Using the framework of the generalized linear models (GLM) introduced  in \cite{nelder1972},  \cite{behzadi2019granger} proposed a general model to detect Granger-causal relations among $p \ge 3$ number of  time series which follow a distribution from the exponential family. The relationship among the response variable and the covariates in a regression is not linear  but defined by a so-called link function $\eeta$, which is a monotone, twice differentiable function and depends on a concrete distribution function from the exponential family.

The heterogeneous graphical Granger model (HGGM) from \cite{behzadi2019granger}, considers time series  $\bx_i$ that follow a distribution from the exponential family\index{exponential distribution} using a canonical parameter $\btheta_i$.  The generic density form for each $\bx_i$ can be written as:
%(McCullagh and Nelder 1989)
\begin{equation}\label{definition_of_p_of_fHGGM}
    p(\bx_i|\bX^{Lag}_{t,d}, \btheta_i) = h(\bx_i)\exp(\bx_i\btheta_i -\eta_i(\btheta_i))
\end{equation}

\noindent where $\btheta_i =\bX^{Lag}_{t,d}(\bbeta_{i}^*)'$, with $\bbeta_{i}^*$ being the optimum,
and $\eta_i$ is a link function corresponding to time series $\bx_i$.
%where $\theta_i = X^{Lag}_{t,d}\beta_{i}'$. 
%The heterogeneous graphical Granger model
The HGGM uses the idea of generalized linear models and  applies them to time series in  the following form

%\vspace*{-0.3cm}
\begin{equation}\label{ourform2}
x_i^t \approx \mu_i^t = \eta_i^t (\bX^{Lag}_{t,d}\bbeta_{i}')= \eta_i^t (\sum_{j=1}^p \sum_{l=1}^d  x_j^{t-l}\beta_j^l) %+ \varepsilon_i^t
\end{equation}

\noindent for $x_i^t$, $i = 1, \dots, p, t = d+1, \dots, n$ each having a probability density  from the exponential family;
$\bmu_i$ denotes the  mean of $\bx_i$ and $var(\bx_i|\bmu_i, \phi_i)=\phi_i v_i(\bmu_i)$ where $\phi_i$ is a dispersion parameter
and $v_i$ is a variance function dependent only on $\bmu_i$; 
$\eta_i^t$ is the t-th coordinate of $\eeta_i$.
The causal inference in (\ref{ourform2})  can be solved as a maximum likelihood  estimate for $\bbeta_i$

for a given lag $d>0$, $\lambda >0$,  and all $t=d+1, \dots, n$ %and $x_i$ having Poisson distribution, $i = 1, \dots, p$
 with
%$R({\bbeta}_i) $ to be the
added 
adaptive lasso penalty function (see \cite{behzadi2019granger}). 
%The first two summands
%in (\ref{minproblem2}) correspond to 
% the maximum likelihood estimate (MLE) in
%the GLM. 
%As considered in the literature, also here we will consider the case when $\lambda_i:=\lambda$ for  all $i=1, \dots, p$.
%\begin{definition}\label{def3}
Similarly, one can say that \emph{the time series $\bx_j$ Granger--causes  time series $\bx_i$ for a given lag $d$, and denote 
$\bx_j \to \bx_i$, for $i, j = 1, \dots, p$ if and only if at least one of the $d$ coefficients in $j-th$ row 
of $\hat{\bbeta_i}$ of the penalized  solution
%of (\ref{minproblem2}) 
is non-zero}, see \cite{behzadi2019granger}.
%\end{definition}

The idea of the HMML method for estimation of $\bbeta_i$ coefficients is the following: it replaces the solution via $p$ penalized linear regression problems by formulating the feature selection as a combinatorial optimization problem, as it was done in  \cite{hlavackova2020} for  the multivariate Granger causal model with  time series from the exponential family. It uses the information-theoretic criterion "minimum message length" (MML), introduced by \cite{wallace2005statistical}  for  general inference problems, to determine causal connections in the model, improving the results especially for "short"\footnote{The length of a short time series is of the order of at most hundreds of the  number of involved time series.} time series. 
The MML principle is an information-theoretical formulation of Occam’s razor: Even
when models have a comparable goodness-of-fit to the observed data, the one generating
the shortest overall message is more likely to be correct (where the message consists of
a statement of the model, followed by a statement of data encoded concisely using that
model). The statistical version of the MML principle constructs a description in terms of probability functions and some prior knowledge of the parameter vector. MML seeks the model that minimizes this trade-off between model complexity and model
capability.  In the type of MML considered in \cite{hlavackova2020} and in this study and application, the parameter space $\btheta$ for the statistical model
$p(.|\btheta)$ is decomposed into a countable number of subsets and associated code words for
members of these subsets.
The parameter $\btheta$ in the MML criterion corresponds to  the maximum likelihood estimates of the regression coefficients $\bbeta_i$ and the dispersion coefficient of the target time series.  
Each regression problem  for $i=1, \dots, p$  is expressed via incorporation of a subset of indices of regressor variables, 
denoted by $\bgamma_i \subset \{1,..,p\}$ and $k_i = |\bgamma_i|$ into (\ref{ourform2})

\begin{equation}
x_i^t  = \eta_i^t (\bX^{Lag}_{t,d}(\bgamma_i)\bbeta_{i}'(\bgamma_i))= \eta_i^t (\sum_{j=1}^{k_i} \sum_{l=1}^d  x_j^{t-l}\beta_j^l)
%+
\end{equation}
where $\bX^{Lag}_{t,d}(\bgamma_i)$ is the design matrix with regressors only from $\bgamma_i$  and $\bbeta_{i}'(\bgamma_i)$ are their regression coefficients.
The best structure of $\bgamma_i$ in the sense of MML principle is determined either by a genetic or exhaustive search algorithm, for more details see \cite{hlavackova2020}. 
Similarly as in the Gaussian case (Eq.~\ref{VAR_for_Granger}), the  time lag (i.e. the model order) of the target variable in HMML can be determined by expert knowledge or by  information theoretic criteria.

%Since HMML is an instance of GLM models, the consequences about collinear or almost collinear time series hold also 
%for HMML.
%Collinearity does not violate any assumptions of GLMs, unless there is a perfect collinearity.
%\\
For the equations and criterion to compute the causal values explicitly we refer to \cite{hlavackova2020}. We use the MML criterion only for the target variable wind speed (one $i$).  As some climatological  processes are better fitted by exponential distributions than by a Gaussian one, 
 HMML can be beneficial to inference on our data set. 
As demonstrated in synthetic and real experiments in the same publication, HMML significantly improved the causal inference precision of those   in \cite{behzadi2019granger} especially for short time series. This is our case,   as we work with short time series.

\vspace{0.3cm}

\noindent  {\large \textbf{Computation of $RSS_1$ and $RSS_2$ from Eq.~(\ref{granger-moderator2}) and Eq.~(\ref{granger-moderator})}}

\vspace{0.3cm}

\noindent 1. For $RSS_1$: Compute $\hat{\gamma_0}, \hat{\gamma_1}$ for regression 
Eq.~(\ref{granger-moderator2}) 
by maximum likelihood function with the distribution  found by statistical fitting of $y$ in interval $I_2$ before  HMML is applied.
%In Matlab it can be done by this function:
%\url{https://de.mathworks.com/help/stats/mle.html#d126e759313}
%python function
% e.g.
%\url{https://docs.scipy.org/doc/scipy/reference/generated/scipy.stats.fit.html}
2. For $RSS_2$: Compute $\hat{\gamma_0}, \hat{\gamma_1}, \hat{\gamma_2}$ for regression Eq.~(\ref{granger-moderator}) by maximum likelihood function with the distribution  found by statistical fitting of $y$ in interval $I_2$  before  HMML is applied.

\vspace{0.3cm}

\noindent  {\large \textbf{Step 13 of the Cause-Trigger Algorithm }}

\vspace{0.3cm}

Denote $r = n-d$,  the size of vector $V$.
We say that the variable  $x_s$  moderates  
variable $V$ with respect to the effect $y$  on interval  $I_2$ if regression (\ref{granger-moderator}) is 
 statistically significantly  better than regression
(\ref{granger-moderator2}).
This will be decided by the  statistical F-test, where

\begin{itemize}
\item[1.]
 The  null hypothesis $H_0$: $x_s$ does not  moderate  $V$ with respect  to the effect $y$ in $I_2$
 is supported \\
 if $\gamma_2 = 0$, reducing Eq. (\ref{granger-moderator}) to (\ref{granger-moderator2}).
\item[2.] The F-statistics is 
\begin{equation}\label{GS2}
S = \frac{(RSS_1 -RSS_2)/1}{RSS_2/(r-3)}
\end{equation}
where $RSS_1$ is the residual sum of squares corresponding  to 
(\ref{granger-moderator2}),
$RSS_2$ is the residual sum of squares corresponding to regression (\ref{granger-moderator}), and
  $r$ length of the time series in $I_2$.
\end{itemize}

\begin{enumerate}
\item[3.] Calculate the F-statistic S from equation (\ref{GS2}).
\item[4.] %For a one-sided test, the null hypothesis is rejected when the test statistic is greater than the tabled  critical value. 
Reject the null hypothesis $H_0$ that  $\gamma_{2} = 0$ if the F-statistic % (Step 5) 
is greater than the  critical F-value, otherwise accept $H_0$.
\end{enumerate}

\section{Description of possible  influence mechanisms 
of selected variables on cyclogenesis}\label{influence_mechanisms}

Based on the literature,  the following variables are assumed that they can be causal or triggering during cyclogenesis. 
\begin{enumerate}
    \item \textbf{Divergence (d)}  [$s^{-1}$]: 
    Upper-level divergence, in combination with lower-level convergence, enhances upward motion and supports convective clustering, an essential ingredient for tropical cyclone formation, see \cite{Frank:1977, DavisBosart:2003}.

    \item \textbf{Geopotential (z)} [$m^2s^{-2}$]: 
    Geopotential height patterns at mid-levels reflect large-scale synoptic support for cyclogenesis. Lower heights can indicate  pre-existing disturbances, see \cite{frank1977structure}.

    \item \textbf{Ozone mass mixing ratio (o$_3$)} [$kg kg^{-1}$]: 
    Ozone is not traditionally included in tropical cyclone genesis indices but has been linked to stratosphere–troposphere exchange (STE), which can modify thermal structure and stability. Its presence in the mid-troposphere can signal dynamical intrusions that influence cyclogenesis, see \cite{HighwoodHoskins:1998, SprengerWernli:2003}.

    \item \textbf{Potential vorticity (pv)} [$K m^2 kg^{-1}s^{-1}$]:
    Potential vorticity anomalies, especially in the lower and mid-troposphere, are well-documented precursors of tropical cyclogenesis. Elevated PV can enhance vertical coupling and promote cyclonic development, see \cite{MontgomeryEnagonio:1998, ToryMontgomery2006}.

    \item \textbf{Relative humidity (r)} [\%]: 
    Mid-level moisture is a well-established environmental control on tropical cyclogenesis. High humidity at 700 hPa reduces convective inhibition and supports sustained deep convection, see  \cite{marenco2018, Emanuel:2005}.

    \item \textbf{Vertical velocity (w)} [$Pa s^{-1}$]: 
    Vertical motion, particularly strong upward motion, is essential for initiating and maintaining convection. It is often used in composite indices of cyclone potential, see \cite{Zehr:1992, tory2013importance}.

    \item \textbf{Temperature (t)} [$K$]: 
    Thermal gradients influence static stability and convective available potential energy (CAPE). Warm mid-level temperatures can suppress convection, while cooler air aloft promotes instability, see \cite{Emanuel:1986}.

    \item \textbf{Wind speed (ws)} [$ms^{-1}$]: 
    Wind speed, particularly in the lower and mid-troposphere, is directly linked to surface convergence and the organization of vorticity. Strong low-level winds can enhance cyclonic circulation and promote, see vertical stretching \cite{Gray:1979}.

    \item \textbf{Wind direction (wd)} [$^\circ$]: 
    Wind direction is physically linked to convergence, rotational flow, and upper-level ventilation, see \cite{TangEmanuel2012}.

    \item \textbf{Sine of the wind direction (sin(wd))}  [scalar value]:
    The sine of wind direction serves as a directional proxy, reflecting alignment or curvature of the flow—relevant in detecting early cyclonic rotation, see \cite{RitchieHolland:1999}.

\end{enumerate}

\end{document}